\documentclass[a4paper,conference]{IEEEtran}
\usepackage[font=small,labelfont=bf]{caption}
\usepackage{ifpdf}

\usepackage{cite}

\usepackage{rotating}
\usepackage{graphicx}
\usepackage{sidecap}
\usepackage{amsmath}
\usepackage{algorithmic}

\usepackage{array}
\usepackage{fixltx2e}

\usepackage{stfloats}
\usepackage{graphics}
\usepackage{booktabs}

\usepackage{url}

\renewcommand{\thefootnote}{\fnsymbol{footnote}}

\begin{document}
%
\title{SS-CAM: Smoothed Score-CAM for Sharper Visual Feature Localization}

\author{\IEEEauthorblockN{Haofan Wang*}
\IEEEauthorblockN{Electrical and Computer Engineering}
\IEEEauthorblockN{Carnegie Mellon University}
\IEEEauthorblockA{
haofanw@andrew.cmu.edu}
\and \IEEEauthorblockN{Rakshit Naidu* and Joy Michael}
\IEEEauthorblockN{Computer Science and Engineering}
\IEEEauthorblockN{Manipal Institute of Technology}
\IEEEauthorblockA{
nemakallu.rakshit@learner.manipal.edu and\\
manda.joy@learner.manipal.edu}
\and
\IEEEauthorblockN{Soumya Snigdha Kundu
}
\IEEEauthorblockA{Computer Science and Engineering\\
SRM Institute \\
of Science and Technology\\
sk7610@srmist.edu.in}
}

%


\maketitle

\begin{abstract}
Interpretation of the underlying mechanisms of Deep Convolutional Neural Networks has become an important aspect of research in the field of deep learning due to their applications in high-risk environments. To explain these black-box architectures there have been many methods applied so the internal decisions can be analyzed and understood. In this paper, built on the top of Score-CAM, we introduce an enhanced visual explanation in terms of visual sharpness called SS-CAM, which produces centralized localization of object features within an image through a smooth operation. We evaluate our method on the ILSVRC 2012 Validation dataset, which outperforms Score-CAM on both faithfulness and localization tasks.
\end{abstract}


%
\IEEEpeerreviewmaketitle

\let\thefootnote\relax\footnote{*Equal Contribution}

\section{Introduction}
The primary solution to vision problems in the past years have been Convolutional Neural Networks (CNNs). There have been sufficient advancements in its architectures\cite{Resnet18-15}\cite{tan2019efficientnet} to cope with complex problems such as image captioning\cite{8803108}, image classification\cite{classification19}, semantic segmentation\cite{SemanticSeg18} and many other problems\cite{ren2015faster, wang2019hybrid, hu2015face}. Despite their progressive nature of solving major vision tasks, they still act like black box architectures and interpreting these models has become a difficult task. As these architectures are applied to highly sensitive environments such as medicine, finance and autonomous transportation facilities, one has to know the construct for the basis of the architecture's findings to prevent any type of irreparable damage. The task continues to become increasingly difficult as the complexity of the architecture increases. 

To avoid instances where sensitive issues are hampered, these unexplainable architectures need to be justified with visual reasoning to explain the decision that they make.
Not only can the visual information help in identifying the pit-falls of the architectures but also help in generating insights to improve on the model as well. Identifying the pit-falls can help in debugging the network. Models generally tend to mis-classify while performing the above tasks but it takes a lot of time and effort to understand why exactly the model is making such mistakes. The visual information can help in pin-pointing the mistakes of the model. Additional insights can be those related to improving the model in general to generate better localization, faithfulness and gaining human trust.  They can also help in solving the issues explained in \cite{pmlr-v81-buolamwini18a} which focuses on the racial and gender biases involved in non-inclusive datasets which has become a raising concern for many developers. Any architecture which fails to deal with racism and gender bias will not be acceptable on any front.

\cite{zeiler14} by the means of Deconvolution and \cite{Springenberg2015StrivingFS} by the means of Guided Backpropagation were the first ones to address the issue. But recent progress came to notice with the culmination of attribution maps by giving importance to the weights of each region in the image before the final Pooling layer.  Gradient-based visualisation, Perturbation-based visualisation and Class Activation Mappings\cite{CAM15} are the 3 types of maps which help in developing the visual aid. Gradient-based approaches generate saliency map using the derivative of the target class score to the input image through backpropagation \cite{Springenberg2015StrivingFS}. Perturbing \cite{LIME16, lundberg2017unified} is also a commonly adopted technique to find the region of interest in the image, which works by masking some specific region in the input. The region that causes the largest drop on target class is regarded as the important region. Other works\cite{wang2020smoothed} add regularizers to make these attribution maps more robust and some packages\cite{yang2019xdeep} have been developed

Our work builds on CAM-based approaches\cite{CAM15, Gradcam17, GradCAM++18}, which obtain attribution maps by a linear combination of the weights and the activation maps. Recent CAM-based approaches all generalize the original CAM and are not limited to CNNs with a special global pooling layer. They can be divided into two branches, one is gradient-based CAMs\cite{Gradcam17, GradCAM++18}, which represent the linear weights corresponding to internal activation maps by gradient information. The other is gradient-free CAMs\cite{wang2020score, AblationCAM20} which capture the importance of each activation map by the target score in forward propagation. Although Score-CAM\cite{wang2020score} has achieved good performance on both visual comparison and fairness evaluation, its localization result is coarse which leads to certain instances of non-interpretability.


To achieve better visual feature localizing, in this paper, we propose a novel method called SS-CAM, built on top of Score-CAM\cite{wang2020score}, for enhancing object understanding and providing better post-hoc explanations about the centralized, target object in the image using Smoothing. We evaluate our approach on ILSVRC 2012 dataset on the basis of 4 major metrics. Our contributions can be summarized as below:

\begin{itemize}
    \item We introduce an enhanced visual feature localization method SS-CAM, which combine Score-CAM with an extra smooth operation and leads to a visually sharper attribution map.
    
    \item Two types of smoothing are introduced, we quantitatively evaluate the differences in smoothing for different positions.
    
    \item We achieve better visual performance than previous CAM-based methods. We quantitatively evaluate on recognition, localization and human trust tasks and show that SS-CAM better localizes decision-related features.
\end{itemize}



\section{Related Work}

A Gradient-based approach generates a saliency map using the derivative of the target class score to the input image through backpropagation\cite{Springenberg2015StrivingFS}. 

\textbf{SmoothGrad \cite{Smooth17}}: 
SmoothGrad aims to reduce the visual noise by adding sufficient noise to an image and generating similar images. Then the average of the resulting sensitivity maps of each of these sampled, similar images is taken. Training with the noise and inferring the decision of the model with the noise seems to have a better effect at yielding the best results. Visually-sharp gradient-based sensitivity maps are created by taking random samples in the neighbourhood of $x$ and averaging the resulting maps. It is indicated by: 
    \[ \widehat {M_{c}}\left( x\right) =\dfrac {1}{n}\sum         ^{n}_{1}M_{c}\left( x+\mu \left( 0,\sigma ^{2}\right) \right) \]

    where ${n}$ refers to the number of samples and $\mu \left( 0,\sigma ^{2}\right)$ refers to the Gaussian noise with a standard deviation $\sigma$ that is added to an image input ${x}$. 

    


\textbf{CAM \cite{CAM15}:}
    CNNs utilise targeted and distinguishable image regions for identifying particular categories by linearly combining weights with the activation maps. These categories are known as Class Activation Maps (CAMs). The regions that are highlighted as the predicted class score values are mapped back to the previous convolutional layer. It is denoted by :
    
    \begin{equation}
    L^{c}_{CAM}=ReLU\left( \sum _{k}\alpha ^{c}_{k}A^{k}_{l-1}\right)
    \end{equation}
    $where$
    \begin{equation}
    \alpha^{c}_{k} = w^{c}_{l,l+1}[k]
    \end{equation}
    $\alpha^{c}_{k} = w^{c}_{l,l+1}[k]$ refers to the weight of the k-th neuron after the pooling operation.
    
     The generalisation of CAMs take place with Grad-CAM\cite{Gradcam17}.
     To obtain fine-grained visualizations, Grad-CAM activation maps are revamped by multiplying with the visualization obtained via Guided Backpropagation. This is better known as Guided Grad-CAM.
     
\textbf{Grad-CAM \cite{Gradcam17}:} This technique utilizes local gradient to represent linear weight, and can be applied to any Average Pooling-based CNN architectures without the re-training process. It is denoted by :
    \begin{equation}
    L^{c}_{Grad-CAM}=ReLU\left( \sum _{k}\alpha ^{c}_{k}A^{k}_{l}\right)
    \end{equation}
    $where$
    \begin{equation}
    \alpha^{c}_{k}  = GAP\left(\frac{\partial Y^{c}}{\partial A^{k}_{l}} \right)
    \end{equation}

GAP(.) refers to the Global Average Pooling operation.

     However, this approach has some limitations. It fails to provide worthy explanations in highly confident scenarios due to gradient saturation and fails to highlight multiple occurrences of the same object in a single image. Also, the heatmaps generated with Grad-CAM do not capture the entire object completely as the gradients tends to be noisy, which is required to gain better recognition of the object in the image. 
     
     This technique was further improved with Grad-CAM++\cite{GradCAM++18} which gives better results by localizing the entire object in image, instead of bits and pieces of it. 
However, as the original CAM is restricted to CNNs having the GAP layer, they do not generate feasible explanations on a wide array of CNNs. As the output layer is a non-linear function, gradient-based CAMs tend to diminish the backpropagating gradients which cause gradient saturation thereby making it difficult to provide concrete explanations.
To solve these problems, Score-CAM\cite{wang2020score} bridges the gap between perturbation-based and CAM-based method by being the first gradient-free generalization CAM.

\textbf{Score-CAM \cite{wang2020score}:}
Score-CAM obtains the weights of each activation map in its forward passing score on the target class. The final result is a linear combination of both the weights and the activation maps. In Score-CAM, $L^{c}$ is denoted as follows: 

\begin{equation}
L^{c}_{Score-CAM}=ReLU\left(\sum_{k}\alpha^{c}_{k}A^{k}_{l}\right)  
\end{equation}
$where$
\begin{equation}
    \alpha ^{c}_{k} = C(A^{K})
\end{equation}
where $c$ is the class of interest, $l$ is the convolution layer and $C(.)$ is the Channel-wise Increase of Confidence(CIC) score for the activation map $A^{k}_{l}$.


\section{Proposed Approach}

\begin{figure*}[t]
\centering
    \includegraphics[width=13.2cm]{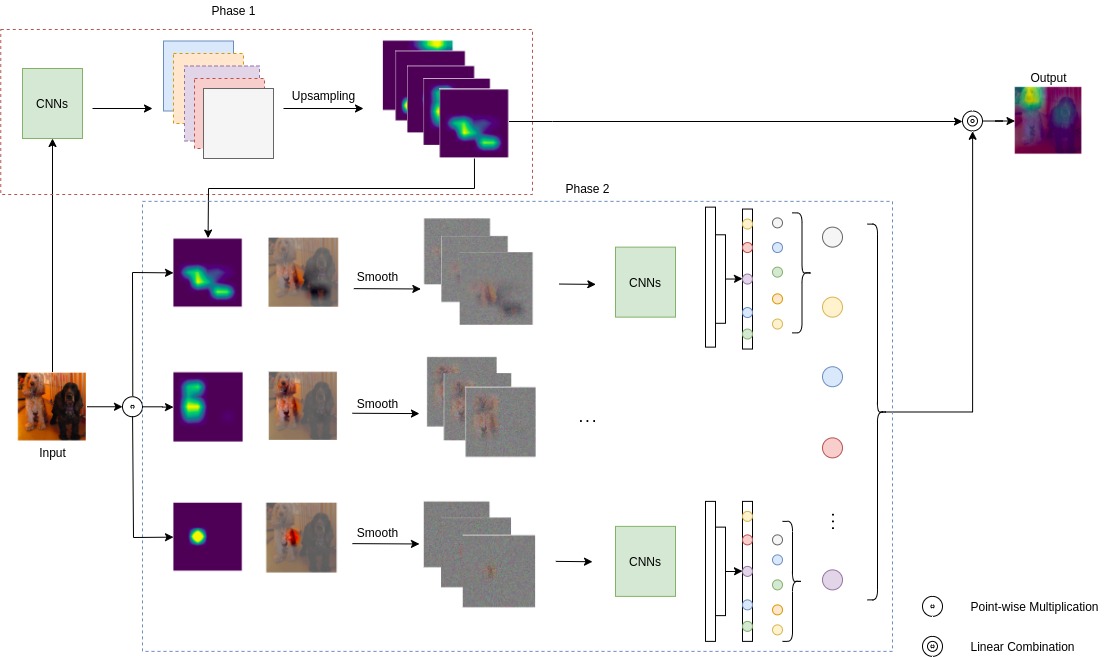}
    \caption{Pipeline of the proposed SS-CAM appraoch. The saliency map is produced by the linear combination of the average score after the smoothing operation and the upsampled activation maps. The average score is obtained from these noisy images and utilized as the final score to provide enhanced feature localizations.}
    \label{fig:SS-CAM_pipeline}
\end{figure*}

Detecting model misbehavious and a good post-hoc visualization were the primary applications of Score-CAM \cite{wang2020score}. With SS-CAM, we generate better visualisations with higher faithfulness evaluation which leads one to analyse the model with a better intuition. We borrow the idea from SmoothGrad\cite{Smooth17} and integrate the smooth into the pipeline of Score-CAM to generate "smoother" feature localization. The higher "smoothness" helps one perform a higher level of analysis on the architecture. In this paper, we evaluate two ways to smooth. The pipeline for the proposed method is shown in Figure ~\ref{fig:SS-CAM_pipeline}. Before diving into the details of techniques, we define our motivation to pursue this topic of interest and why we address it.

\begin{figure}[ht]
\centering
\begin{tabular}{c c c}
\text{Input Image}  &  \text{Score-CAM} & \text{SS-CAM(2)}\\
\includegraphics[width=2cm]{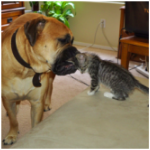} & \includegraphics[width=2cm]{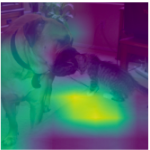} & \includegraphics[width=2cm]{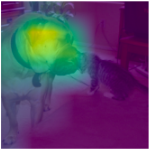}

\end{tabular}

\caption{Depicts the Imagenet Label: Bull Mastiff, and the outputs of Score-CAM and SS-CAM(2). As you can see the Score-CAM output is quite noisy by itself (probably due to ReLU non-linearity), this is what encouraged us to tackle this issue.}
\label{fig:SS-CAM_comp_with_score}
 
\end{figure}

We choose to investigate Smoothing in Score-CAM to generate more feasible explanations. As one can see the Score-CAM map (Figure \ref{fig:SS-CAM_comp_with_score}) clearly doesn't highlight regions that aren't localized to the target object (here, Bull Mastiff). This probably due to the noise generated because of the ReLU non-linearity. The map also spreads all over which clearly makes it highly unreliable to convery any type of result but SS-CAM generates a concentrated heat-map which completely focuses on the object. Such Concentrated heat maps can substantially increase the chance of finding faults with the predictions. If a model is mis-classifying, the mis-classified heat map will appropriately show the part of the image which is being represented and help in debugging the model. We now proceed to explain our approaches starting by measuring the importance of each activation map with CIC:
\begin{figure*}[h]
\begingroup
\setlength{\tabcolsep}{5pt}
\begin{tabular}{c c c c c c c}
\text{} & \text{} & \text{} & \text{Smooth} \\
{\hskip 0.01in}
\text{Input Image}&
{\hskip 0.17in}
\text{Grad-CAM}& 
{\hskip 0.129in}
\text{Grad-CAM++}&
{\hskip -0.011in}
\text{{\hskip 0.000in} Grad-CAM++}&
{\hskip 0.1in}
\text{Score-CAM}& 
{\hskip 0.16in}
\text{SS-CAM(1)}& 
{\hskip 0.14in}
\text{SS-CAM(2)}\\
\end{tabular}
\endgroup
    \includegraphics[width=18cm]{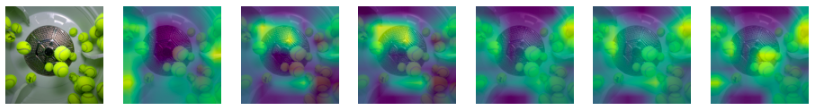}
    \includegraphics[width=18cm]{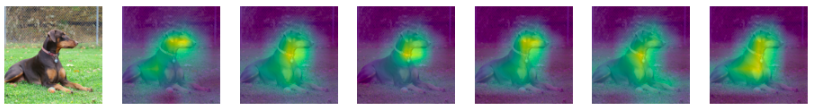}
    \includegraphics[width=18cm]{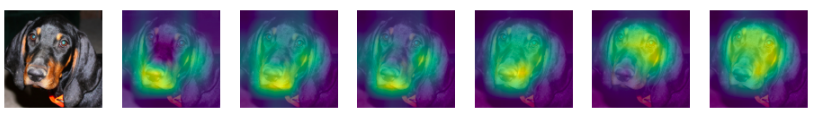}
    \caption{Depicts the Imagenet Labels (Row-wise): Tennis ball, Doberman and Black-and-tan Coonhound. 
    This figure is used for a Visual Comparison of our approach with the other existing approaches.}
    \label{fig:SS-CAM_visual_comparison}
\end{figure*}

\begin{table*}[ht]
\caption{AUC scores of the Deletion curve(the lower the better) and Insertion curve(the higher the better) for the Black-and-tan Coonhound saliency maps as generated in Figure \ref{fig:SS-CAM_visual_comparison}.}
\small
\label{table1}
\begin{center}
\resizebox{12cm}{!} {
\begin{tabular}{lcccccr}
\toprule
Metrics & Grad-CAM & Grad-CAM++ & Smooth & Score-CAM & SS-CAM(1) & SS-CAM(2) \\
 &  &  &  Grad-CAM++ \\
\midrule
Deletion(\%) & 48.08 & 52.53 & 44.41 & \textbf{28.29} & 43.96 & 29.94 \\
Insertion(\%) & 59.85 & 51.68 & 64.91 & 59.41 & \textbf{70.02} & 61.76  \\
\bottomrule
\end{tabular}
}
\end{center}
\end{table*}

\begin{table*}[t]
\caption{Average Drop(the lower the better) and Average Increase In Confidence(the higher the better) across 2000 ILSVRC Validation images with VGG-16.}
\small
\label{table_avg_drop}
\begin{center}
\resizebox{12cm}{!} {
\begin{tabular}{lcccccr}
\toprule
Metrics & Grad-CAM & Grad-CAM++ & Smooth & Score-CAM & SS-CAM(1) & SS-CAM(2) \\
 &  &  &  Grad-CAM++ \\
\midrule
Average Drop(\%) & 89.69 & 89.76 & 90.92 & 90.07 & \textbf{89.01} & 90.17 \\
Average Increase(\%) & 0.65 & 0.40 & 0.45 & 0.60 & 0.65 & 0.65  \\
\bottomrule
\end{tabular}
}
\end{center}
\end{table*}

\textbf{Channel-wise Increase of Confidence(CIC)} : 
\textit{Given a model $f(X)$ that takes an image input $X$, the $k$-th channel of an activation map $A$ of a convolutional layer $l$ in $f$ is given as $A^{k}_{l}$. For a baseline input $X_{b}$ the importance of $A^{k}_{l}$ is denoted as}
\begin{equation}
    C(A^{k}_{l}) = f(X \cdot H^{k}_{l}) - f(X_{b})
\end{equation}
$where$
\begin{equation}
    H^{k}_{l} = s(U(A^{k}_{l}))
\end{equation}
\textit{Here, $U(.)$ is the Upsampling operation of the activation map $A^{k}_{l}$ into the size of the input and $s(.)$ refers to the normalization function (as explained in Equation 15)) so that the elements are within the [0,1] range.}

A few similar approaches to CIC are DeepLIFT\cite{DeepLIFTSGK17} and Average Drop\% and Increase in Confidence from \cite{GradCAM++18}. The significance of activation maps are represented by the gradient information that flows into the final convolutional layer.

\textbf{Smooth on activation map}
We set the number of noised sample images $N$ to be generated by adding Gaussian noise to each activation map $A_{i}$. For each activation map $A_{i}$, $N$ scores are generated and averaged to a final score, which is considered as the importance of the activation map $A_{i}$. 

An activation map $A$ of a convolutional layer $l$ with channel $k$ is given as $A^{k}_{l}$. SS-CAM with smooth on feature space (type1) can be depicted as:
\begin{equation}
L^{c}_{SS-CAM} = ReLU\left( \sum _{k}\alpha ^{c}_{k}A^{k}_{l}\right)
\end{equation}
$where$
\begin{equation}
    \alpha^{c}_{k} = \dfrac{\sum ^{N}_{1}\left(C(M)\right)}{N}
\end{equation}
\textit{Here, $c$ is the class of interest, $l$ is the convolutional layer and $\alpha^{c}_{k}$ is the average Channel-wise score of the activation map $A^{k}_{l}$ which accounts for the scores generated by the noisy maps $A^{k}_{l} + N(0, \sigma)$. $X_{0}$ is the input image and $A^{k}_{l}$ refers to the activation map at layer $l$.
\begin{equation}
    M = \sum ^{N}_{1} (X_{0} * (A^{k}_{l} + N(0,\sigma))) 
    \label{eq:15}
\end{equation}}

\textbf{Smooth on input} : An activation map $A$ of a convolutional layer $l$ with channel $k$ is given as $A^{k}_{l}$. SS-CAM with smooth on input space (type2) can be depicted as:
\begin{equation}
L^{c}_{SS-CAM}=ReLU\left( \sum _{k}\alpha ^{c}_{k}A^{k}_{l}\right)
\end{equation}
$where$
\begin{equation}
    \alpha^{c}_{k} = \dfrac{\sum ^{N}_{1}\left(C(M)\right)}{N}
\end{equation}
\textit{Here, $c$ is the class of interest, $l$ is the convolutional layer and $\alpha^{c}_{k}$ is the Channel-wise average score of the noisy normalized activation map ($A^{k}_{l} + N(0, \sigma)$). $X_{0}$ is the input image and here, $X_{0}*A^{k}_{l}$ refers to the normalized input mask.
\begin{equation}
    M = \sum ^{N}_{1} ((X_{0} * A^{k}_{l}) + N(0,\sigma))
\end{equation}}

\textbf{Normalization}: The binary mask over the normalized map would not be great as our aim is to focus on the spatial region where the object lies and the binary mask would lose sight of the important features. Hence, we employ a similar normalization function as used in \cite{wang2020score} to elevate the features within the specific region. 

The normalization used in the algorithm is given as :

\begin{equation}
s\left(A^{k}_{l}\right) = \dfrac{A^{k}_{l} - min(A^{k}_{l})}{max(A^{k}_{l}) - min(A^{k}_{l})}
\end{equation}


\section{Experiments}

In this section, we explain our experimental setting to assess the proposed method(s). Our setting is similar to that in \cite{GradCAM++18, wang2020score, RISE18}. First, a qualitative comparison of the the outputs of the architectures via a visualization on the ILSVRC 2012 Validation set in A. Second, we evaluate the fairness of the explanations of the architectures on object recognition in B. Next, the bounding box evaluations are performed for the  class-conditional localization of objects in a given image in C using the Energy-based pointing game (introduced in \cite{GradCAM++18}) over 2000 uniformly random selected images from the ILSVRC Validation Set 2012. Lastly, we show a quantitative study of Human Trust evaluation to understand how  our  method  achieves  sufficient  interpretability.

Our comparison extends with namely 4  other  existing  CAM techniques, Grad-CAM, Grad-CAM++, Smooth Grad-CAM++  and  Score-CAM. 
The images are resized with a definite size (224, 224, 3), transformed into the [0,1] range and then, normalized using ImageNet\cite{Imagenet} weights (mean vector : [0.485, 0.456, 0.406] and standard deviation vector [0.229, 0.224, 0.225]). For simplicity, baseline image $X_{b}$ is set to 0.

\subsection{Visual Comparison}

To conduct this experiment, 2000 images were randomly selected from the ILSVRC 2012 Validation set. 
Figure \ref{fig:SS-CAM_visual_comparison} displays a few images that compare our approach(es) with existing CAM approaches namely Grad-CAM, Grad-CAM++, Smooth Grad-CAM++ and Score-CAM. Here, we used $N$ = 35 and $\sigma$ = 2. A further depth to this comparison was added to sub-section D.

\subsection{Faithfulness Evaluations}

The faithfulness evaluations are carried out as depicted in Grad-CAM++\cite{GradCAM++18} for the purpose of Object Recognition. Three metrics called Average Drop, Average Increase In Confidence and Win \% are introduced.

These metrics are evaluated over the pre-trained VGG-16 model for 2000 images randomly selected from the ILSVRC 2012 Validation set. We used $N$ = 10 and $\sigma$ = 2 to conduct this sub-experiment.

Insertion and Deletion Curves are used to calculate the Area Under Curve(AUC) metric to understand how many pixels of the saliency map can either contribute or decrease the scores of the resulting fractioned maps. We adapt a similar setting as introduced in \cite{RISE18}. The Deletion operation shows the ability to destroy the map information pixel-wise. A sharp drop and a lower AUC of the scores generated is indicative of a good explanation.
The Insertion operation determines the ability to reconstruct the saliency map from a given baseline. A sharp increase and a higher AUC of the scores generated is indicative of a good explanation.

\begin{figure*}[h]

\includegraphics[width=0.3059\textwidth]{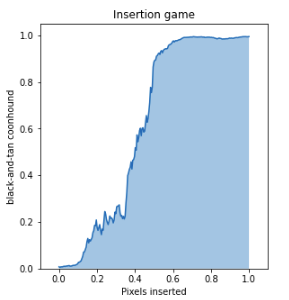}
\includegraphics[width=0.3\textwidth]{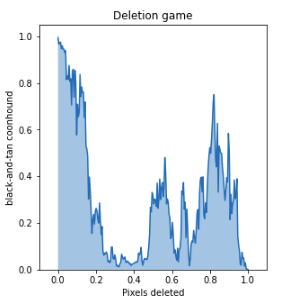}
\centering
\caption{Insertion and Deletion curve of the Black-and-tan Coonhound maps generated above for SS-CAM(2). You can notice a sharp increase midway in the Insertion curve, which indicates that our approach is capable of generating viable explanations.}
\label{fig:SS-CAM_ins_del}
\end{figure*}

\subsubsection{Average Drop \%}

The Average Drop refers to the maximum positive difference in the predictions made by the prediction using the input image and the prediction using the saliency map.  It is given as: 
$\sum ^{N}_{1}\dfrac {\max \left( 0, Y^{c}_{i}-O^{c}_{i}\right) }{Y^{c}_{i}}$ $\times 100$. 
Here, $Y^{c}_{i}$ refers to the prediction score on class $c$ using the input image $i$ and $O^{c}_{i}$ refers to the prediction score on class $c$ using the saliency map produced over the input image $i$.

\subsubsection{Increase in Confidence \%}

The Average Increase in Confidence is denoted as: $ \sum ^{N}_{1}\dfrac {Func\left(Y^{c}_{i} < O^{c}_{i}\right)}{N} $ $\times$ 100 where $Func$ refers to a boolean function which returns $1$ if the condition inside the brackets is true, else the function returns $0$. The symbols are referred to as shown in the above experiment for Average Drop.

\subsubsection{Win \%}

The Win percentage refers to the decrease in the model's confidence for an explanation map generated by SS-CAM. This metric is compared to the confidence generated by Score-CAM\cite{wang2020score} maps with both SS-CAM(1) and SS-CAM(2) maps. SS-CAM(1) when compared with Score-CAM shows 49.8\% of the scores towards its favour while SS-CAM(2) displays a Win \% of 47.8\% scaled on its behalf when compared with the scores by Score-CAM. We use $N$=10 and $\sigma$=2 for this experiment. 

Table \ref{table_avg_drop} depicts moderate to similar performance in regards to average drop and increase in confidence. While Table \ref{table1} depicts SS-CAM(2) just slightly falling short of Score-CAM in terms of the deletion curve, SS-CAM(1) performs quite well with a 5 \% increase in the insertion curve.

\subsection{Localization Evaluations}

Bounding box evaluations are accomplished in this section. We employ the similar metric, as specified in Score-CAM, called the Energy-based pointing game. Here, the amount of energy of the saliency map is calculated by finding out how much of the saliency map falls inside the bounding box. Specifically, the input is binarized with the interior of the bounding box marked as 1 and the region outside the bounding box as 0. Then, this input is multiplied with the generated saliency map and summed over to calculate the proportion ratio, which given as - $Proportion = \dfrac{\sum L^{c}_{(i,j)\in bbox}}{\sum L^{c}_{(i,j)\in bbox} + \sum L^{c}_{(i,j)\not\in bbox}}$.
Three pretrained models namely, VGG-16\cite{simonyanvgg162014}, ResNet-18(Residual Network with 18-layers)\cite{Resnet18-15} and SqueezeNet1.0\cite{Squeeze16} model are used to conduct the Energy-based pointing game on the 2000 randomly chosen images from the ILSVRC 2012 Validation set. We used $N$ = 10 and $\sigma$ = 2 to conduct this sub-experiment. Though the margins are small Table \ref{table_loc} portrays a better localization evaluation for 2 of the 3 architectures.

\begin{table*}[t]
\caption{Localization Evaluations using Energy-based Pointing Game (Higher the better)}
\small
\label{table_loc}
\begin{center}
\resizebox{12cm}{!} {
\begin{tabular}{lcccccr}
\toprule
Metrics & Grad-CAM & Grad-CAM++ & Smooth & Score-CAM & SS-CAM(1) & SS-CAM(2) \\
 &  &  &  Grad-CAM++ \\
\midrule
\textbf{VGG-16} \\
Proportion(\%) & 39.95 & 40.16 & 40.13 & 40.10 & \textbf{40.27} & 39.70 \\
\textbf{ResNet18} \\
Proportion(\%) & 40.90 & 40.85 & 40.86 & 40.76 & \textbf{41.55} & 40.51 \\
\textbf{SqueezeNet1.0} \\
Proportion(\%) & 39.47 & 39.30 & 39.25 & \textbf{39.56} & 37.07 & \textbf{39.56} \\
\bottomrule
\end{tabular}
}
\end{center}
\end{table*}

\begin{table*}[ht]
\caption{Human Trust Evaluations using the visualizations generated by VGG-16 (Higher the better)}
\small
\label{table_hte}
\begin{center}
\resizebox{14cm}{!} {
\begin{tabular}{lccccccr}
\toprule
Metrics & Grad-CAM & Grad-CAM++ & Smooth & Score-CAM & SS-CAM(1) & SS-CAM(2) & Not Sure \\
 &  &  &  Grad-CAM++ \\
\midrule
Responses(\%) & 9.76 & 14.44 & 16.84 & 14.92 & 14.60 & \textbf{20.60} & 8.84 \\
\bottomrule
\end{tabular}
}
\end{center}
\end{table*}

\subsection{Evaluating Human Trust}

In this experiment, we evaluate on human trust or interpretability of the produced visual explanations on the basis of blind reviews. We generate explanation maps for 50 images of the 5 randomly selected classes out of 1000 classes from the ILSVRC 2012 Validation dataset\cite{Imagenet}, totalling to 250 images. The Grad-CAM maps are treated as the base-line for the visualization maps generated. These maps, along with the input image, are shown to 10 subjects (who have no knowledge in this field or Deep Learning) and they are testified to answer which map gives more insight to the captured object, thereby invoking more trust and higher interpretability. A pretrained VGG-16 model was used to create the explanation maps. We choose 5 classes at random as this would give an equal probability of all the classes getting to be picked from the 1000 ImageNet classes, hence, removing any bias.

For each image, six explanation maps were generated namely, Grad-CAM, Grad-CAM++, Smooth Grad-CAM++, Score-CAM, SS-CAM(1) and SS-CAM(2). The subjects were asked to choose which maps demonstrated a better visual explanation of the object in the image (without having any knowledge of which map corresponded to which method). The responses were normalized so that the total score for each image would be equal to 1. These normalized scores were then added and so that the maximum score achievable would be 250 for each subject. The total number of responses recorded are 10*250 = 2500 responses. A "Not sure" option was also provided for the subjects if they were unable to distinguish the visual maps. Table IV shows the corresponding score achieved using this evaluation metric. We used $N$ = 35 and $\sigma$ = 2 to conduct this experiment. The values presented in Table \ref{table_hte} clearly poses the higher response to the SS-CAM(2) approach which leads to the conclusion of higher human interpretability and trust.


\section{Conclusion}
Our proposed method significantly enhances the localization
of the features of the target class in an image, thereby explaining the images profoundly. In this aspect, our method
fairs well above the existing CAM approaches as evaluated
using the Energy-based Pointing game (Table \ref{table_loc}). The higher AUC scores presented in (Table \ref{table1}) further supports the capabilities of the techniques in terms of faithfulness evaluations and (Table \ref{table_hte}) shows a clear superiority in terms of human trust evaluations. The Insertion curve(Figure \ref{fig:SS-CAM_ins_del}) also suggests that SS-CAM has great potentials due to a significant increase in the score as the pixels get inserted.

The Gaussian noise with the paramater $\sigma$\cite{Smooth17, Goh2020UnderstandingIG} is used to smoothen the input space of the attribution maps and construct visually sharp attribution maps. If the parameter $\sigma$ is played around with, it has been found that if $\sigma$ is low, the score generated is quite "noisy" because the resulting attribution maps that tend to be too noisy and if the parameter is high, the attribution map generated is expected to be much more generalized across different classes which would mean that the maps would be rendered useless and irrelevant. Hence, an ideal $\sigma$ is carefully chosen for our evaluations to get the best results.

The limitations of SS-CAM are that it takes a lot of time to process these explanations when compared to Grad-CAM\cite{Gradcam17}. This occurs due to the iteration over each feature map to calculate the scores of $N$ noisy samples during the upsampling stage and averaging them for generating smoother attribution maps. But one should note that the application of such techniques are focused on environments where even a single error can cause great harm and damage. Hence we gave priority to results over the time taken to the obtain the same.


\section{Future Work}
The future prospect is to extend other technologies to CAM and establish a unified framework for each CAM-based method in mathematical expression.

\bibliography{SS-CAM}

\bibliographystyle{IEEEtranS}

\end{document}